\documentclass[sigconf]{acmart}
\renewcommand\footnotetextcopyrightpermission[1]{}
\usepackage{booktabs} 
\usepackage{soul}
\usepackage[markup=default]{changes}
\setdeletedmarkup{\color{red} \st{#1}}
\graphicspath{
    {Figures/}
}
\setcopyright{none}

\begin{document}
\title{Geoseg: A Computer Vision Package for Automatic Building Segmentation and Outline Extraction}

\author{Guangming Wu}
\orcid{0000-0002-2358-5262}
\affiliation{%
  \institution{The University of Tokyo}
  \streetaddress{Center for Spatial Information Science}
  \city{Kashiwa}
  \state{Chiba}
  \country{Japan}
  \postcode{277-8568}
}
\email{huster-wgm@csis.u-tokyo.ac.jp}

\author{Zhiling Guo}
\affiliation{%
  \institution{The University of Tokyo}
  \streetaddress{Center for Spatial Information Science}
  \city{Kashiwa}
  \state{Chiba}
  \country{Japan}
  \postcode{277-8568}
}
\email{guozhilingcc@csis.u-tokyo.ac.jp}

\renewcommand{\shortauthors}{G. Wu \& Z. Guo}

\begin{abstract}
Recently, deep learning algorithms, especially fully convolutional network based methods, are becoming very popular in the field of remote sensing. However, these methods are implemented and evaluated through various datasets and deep learning frameworks. There has not been a package that covers these methods in a unifying manner. In this study, we introduce a computer vision package termed Geoseg that focus on building segmentation and outline extraction. Geoseg implements over nine state-of-the-art models as well as utility scripts needed to conduct model training, logging, evaluating and visualization. The implementation of Geoseg emphasizes unification, simplicity, and flexibility. The performance and computational efficiency of all implemented methods are evaluated by comparison experiment through a unified, high-quality aerial image dataset.
\end{abstract}

%
%

\ccsdesc[300]{Information systems~Geographic information systems}
\ccsdesc[300]{Computing methodologies~Computer vision}
\ccsdesc[300]{Computing methodologies~Image segmentation}

\keywords{Open Source; Computer Vision; Building Segmentation; Outline Extraction; Remote Sensing; Deep Learning;}

\maketitle

\section{Introduction} \label{sec:intro}

Automatic, robust, and accurate image segmentation is a long existing challenge in computer vision. Over the past decades, many supervised or unsupervised methods are proposed to handle this task \cite{sahoo1988survey,ozer2010supervised}. However, due to the limitations of both the quality of dataset and processing algorithm, the precision level of these methods are quite limited \cite{guo2017village}. Recent years, thanks to the rapid development of deep convolutional neural networks (DCNNs) as well as the dramatically increased availability of large-scale datasets, the performances show significant improvement in many image segmentation tasks \cite{lin2014microsoft, everingham2010pascal}.

Differ to ordinary images, because of cost, technical requirement and sensitivity of national defense, it is rather difficult to get very high-resolution (VHR) aerial imagery in the field of remote sensing. And, the lack of large-scale, high-resolution dataset limits the development of accurate building segmentation and outline extraction. Recently, due to rapid evolution of imaging sensors, the availability and accessibility of high-quality remote sensing datasets have increased dramatically \cite{ma2017review, ji2018fully}. On the basis of these datasets, many well-optimized and innovative methods, including different variants of fully convolutional networks(FCNs), have been developed for the purpose of accurate building segmentation \cite{li2018multiple}. Generally, these methods achieve the state-of-the-art accuracy or computational efficiency under corresponding datasets. However, since these methods are trained and evaluated through different datasets, it is hard to have an in-depth comparison of performances of various models. Additionally, although the datasets are open-access, the implemented models or algorithms are usually not revealed in details by the authors.

Facing this problem, we introduce Geoseg (\url{https://github.com/huster-wgm/geoseg}), a computer vision package that is focus on implementing the state-of-the-art methods for automatic and accurate building segmentation and outline extraction. The Geoseg package implements more than 9 FCN-based models including FCNs \cite{long2015fully}, U-Net \cite{ronneberger2015u}, SegNet \cite{badrinarayanan2017segnet}, FPN \cite{lin2017feature}, ResUNet \cite{xu2018building}, MC-FCN \cite{wu2018automatic}, and BR-Net \cite{wu2018boundary}. For in-depth comparison, balanced and unbalanced evaluation metrics, such as precision, recall, overall accuracy, f1-score, Jaccard index or intersection over union (IoU) \cite{real1996probabilistic} and kappa coefficient \cite{stehman1996estimating}, are implemented.

The main contributions of this study are summarized as follows:
\begin{itemize}
  \item We build a computer vision package that implemented several state-of-the-art
  methods (i.e., BR-Net) for building segmentation and outline extraction of very high-resolution aerial imagery;
  \item We have carefully trained and evaluated different models using the same dataset to produce a performance benchmark of various models.
  \item The package is optimized and opened to the public that other researchers or developers can easily adopt for their own researches.
\end{itemize}

The rest of the study is organized as follows: the related work is presented in Section \ref{sec:related}. The benchmark dataset and implementation details of the experiments are described in Section \ref{sec:experimens}. In Section \ref{sec:results&discussion}, the results and discussion of different models are introduced. Conclusions regarding our study are presented in Sections \ref{sec:conclusion}, respectively.

\section{Related work} \label{sec:related}
To assist deep learning researches or applications, there are several deep learning frameworks. According to the compiling mechanism, these frameworks can be categorized into two groups: static and dynamic framework. Static frameworks, such as Caffe \footnote{http://caffe.berkeleyvision.org/} and TensorFlow \footnote{https://www.tensorflow.org/}, construct and compiled completed model before training and updating parameters. For dynamic frameworks, such as Chainer \footnote{https://chainer.org/} and PyTorch \footnote{https://pytorch.org/}, at every iteration, only executed part of the model is compiled. Compared with static frameworks, the dynamic frameworks are less efficient but much flexible.

For different frameworks, there are "Model Zoo" packages that implemented with various pre-trained deep learning models. However, most of the implemented models are focused on methods for image classification. Even for image segmentation packages such as ChainerCV \footnote{https://github.com/chainer/chainercv}, the implemented methods are quite limited, and datasets are not relevant to aerial imagery.

As far as we know, Geoseg is the first computer vision package that implemented with abundant deep learning models for automatic building segmentation and outline extraction.

\section{Experiments} \label{sec:experimens}
\subsection{Benchmark Dataset}
Thanks to the trend of open source, more and more high-quality aerial imagery datasets are available. Among them, a very high-resolution(VHR) aerial image dataset called Aerial Imagery for Roof Segmentation(AIRS) (\url{https://www.airs-dataset.com/}) is published most recently \cite{chen2018aerial}. The spatial resolution of the dataset reaches 0.075 cm. The original orthophotos and corresponding building outlines are provided by Land Information of New Zealand (LINZ) (\url{https://data.linz.govt.nz/layer/53413-nz-building-outlines-pilot/}). For the purpose of accurate roof segmentation, the vectorized building outlines are carefully adjusted to ensure that all building polygons are strictly aligned with their corresponding roofs.

To have a fair comparison of different methods, a study area of AIRS that covers 32 km${^2}$ in Christchurch is chosen \cite{wu2018boundary}.
The study area is evenly divided into two regions: training and testing. For each area, there are 28,786 and 26,747 building objects, respectively. Before experiments, both regions are processed by a sliding window of 224 ${\times}$ 224 pixels to generate image slices (without overlap). After filter out image slices with low building coverage rates from training region, the number of samples in training, validation, and testing data are 27,912 11,952 and 71,688, respectively.

\subsection{Implementation}
\subsubsection{Code Organization}
\begin{figure}[!hbt]
\centering
\includegraphics[width=\columnwidth]{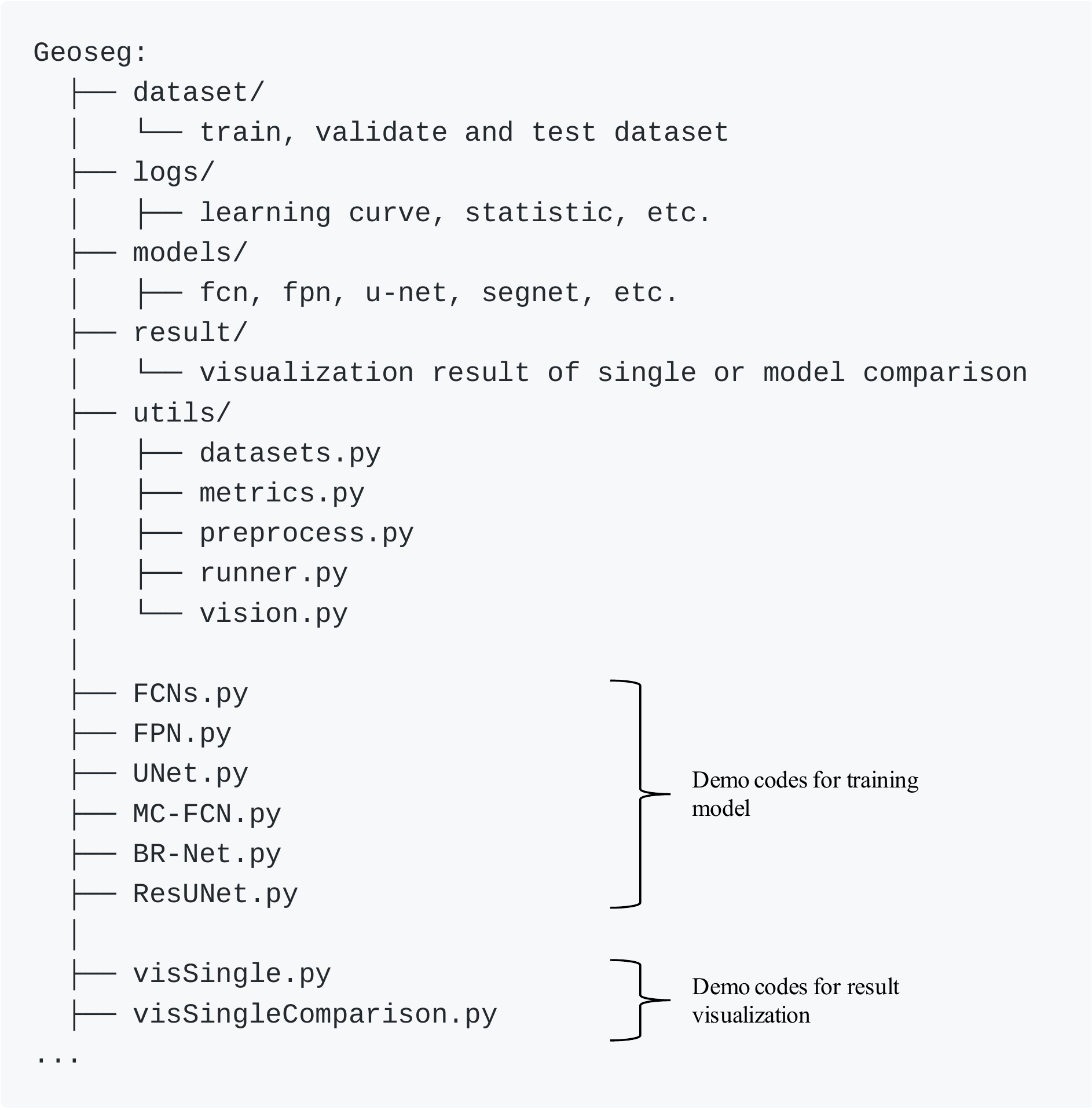}
\caption{The code organization of Geoseg package. The package implements model constructing, training, logging, evaluating and result visualization modules.}
\label{fig:organized}
\end{figure}

Geoseg is built on top of PyTorch with version == 0.3.0 (updating to the latest version is scheduled). The whole package is organized as Figure \ref{fig:organized}. There are 5 sub-directories including dataset/, logs/, models/, result/ and utils/. The dataset/ directory contains all samples for training, validating and testing. The logs/ directory records learning curves, training and validating performance during model iterations. The models/ directory contains scripts implemented with various network architectures of the models. The visualization results are saved in result/ directory. The utils/ directory implements scripts for handling dataset, running instruction, evaluation metrics and visualization tools.

For scripts (e.g., FCNs.py, FPN.py, and UNet.py) at root directory of Geoseg, demo codes for training, logging and evaluating specific models are presented.

For scripts starting with "vis" (e.g., visSingle.py and visSingleComparison.py), demo codes for result visualization of a single model or various models comparison are implemented.

\subsubsection{Models}
In Geoseg, we implemented over 9 FCN-based models according to the reports from original papers. Since the original methods were implemented in various platform and used for various sizes of input, Geoseg introduces few modifications on several models for unification. The details of the implemented models are listed as follows:
\begin{enumerate}
  \item{FCNs}. The classic FCNs method is proposed by Long et al. at 2015. This method innovatively adopts sequential convolutional operations and bilinear upsampling to performance pixel-to-pixel translation. According to fusion and upsampling level of different intermediate layers, the FCNs methods have three variants: FCN32s, FCN16s, and FCN8s.
  \item{U-Net}. The U-Net method is proposed by Ronneberger et al. at 2015. This method adopts multiple skip connections between upper and downer layers.
  \item{FPN}. The FPN method is published on ${\emph{CVPR2017}}$. Similar to U-Net, this method adopts multiple skip connections. Besides, the FPN model generates multi-scale predictions for final output.
  \item{SegNet}. The SegNet method is proposed by Badrinarayanan et al. at 2017. As compared with FCNs, SegNet adopts unpooling which utilizes pooling index of corresponding max-pooling operation to perform upsampling.
  \item{ResUNet}. The ResUNet method adopts the basic structure of U-Net and replaces the convolutional block of VGG-16 \cite{simonyan2014very} with Residual block \cite{he2016deep}. This architecture enhances the representation ability of the model and gains better model performance.
  \item{MC-FCN}. The MC-FCN method is proposed by Wu et al. at 2018. The MC-FCN adopts the U-Net as backend and introduces multi-constraints of corresponding outputs.
  \item{BR-Net}. The BR-Net method is published by ${\emph{Remote Sensing}}$ at 2018. The method utilizes a modified U-Net, which replaces traditional ReLU with LeakyReLU(with ${\alpha=0.1}$), as shared backend. Besides, extra boundary loss is proposed to regulate the model.
\end{enumerate}

Because of the effectiveness of batch normalization(BN) \cite{ioffe2015batch}, advanced models, including FPN, SegNet, ResUNet, MC-FCN, and BR-Net, heavily adopt BN layers after each convolutional operations to increase training speed and prevent bias.

\section{Results and Discussion} \label{sec:results&discussion}
Three FCN variants (FCN8s, FCN18s, and FCN32s), SegNet, U-Net, FPN, ResUNet, MC-FCN, and BR-Net model are adopted as baseline models for comparisons. These models are trained and evaluated utilizing the same dataset and processing platform.

\subsection{Qualitative Result}
\begin{figure}[!hbt]
\centering
\includegraphics[width=\columnwidth]{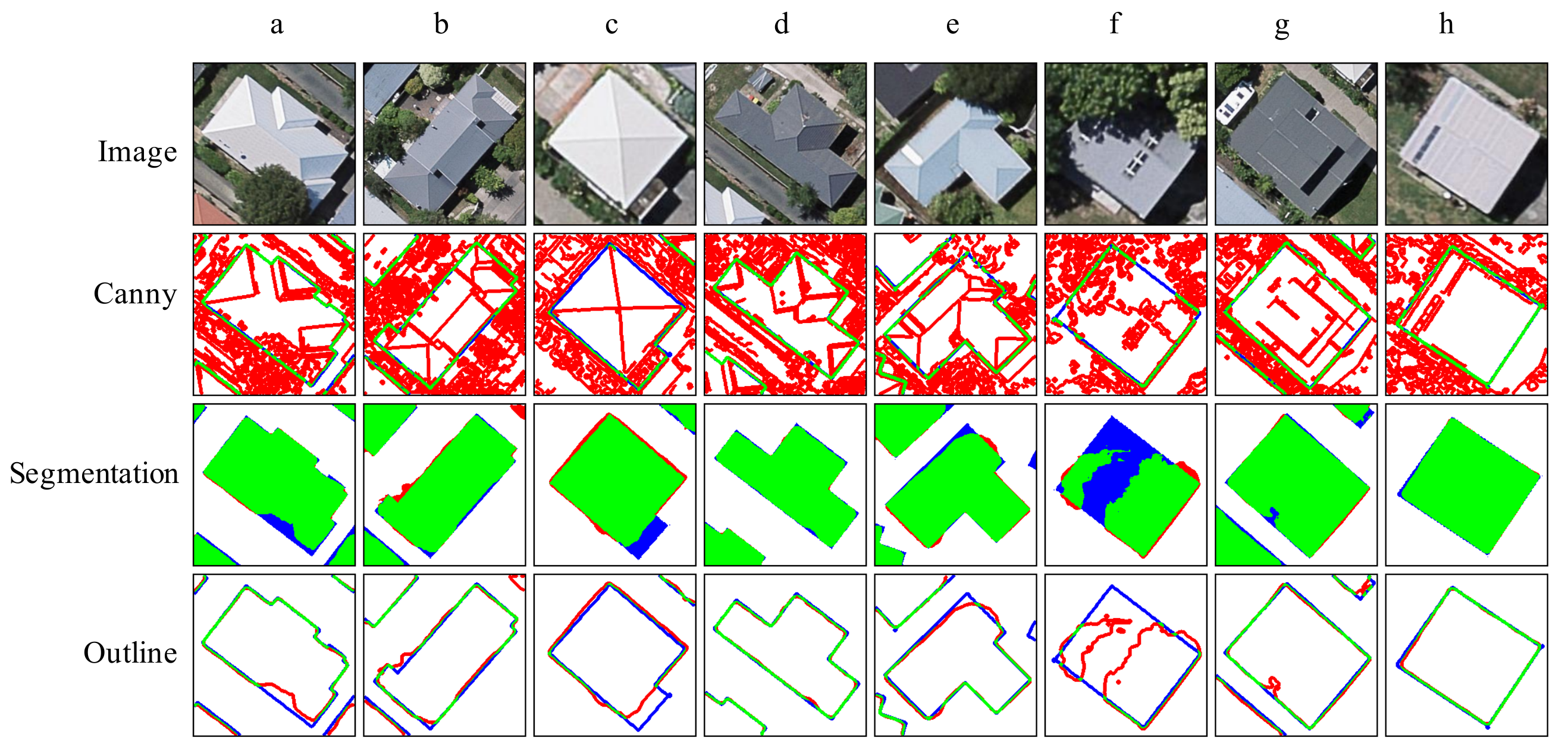}
\caption{Randomly selected eight samples of visualization result. The green, red, blue, and white channels in the results represent true positive, false positive, false negative, and true negative predictions, respectively.}
\label{fig:single}
\end{figure}

Figure \ref{fig:single} presents eight groups of randomly selected visualization results generated by BR-Net. From top to bottom rows, there are original images, extracted edges by Canny, building segmentation and outline extraction from BR-Net model. In general, the extracted outlines through Canny detector contains pretty much noise (see $2^{nd}$ Row). The BR-Net can segment the major part of buildings from most of the selected RGB images (see $3^{rd}$ Row). Building outlines extracted from segmentation results show much fewer false negatives (see $2^{nd}$ Row vs. $4^{th}$ Row).

\subsection{Quantitative Result}
\begin{figure}[!hbt]
\centering
\includegraphics[width=\columnwidth]{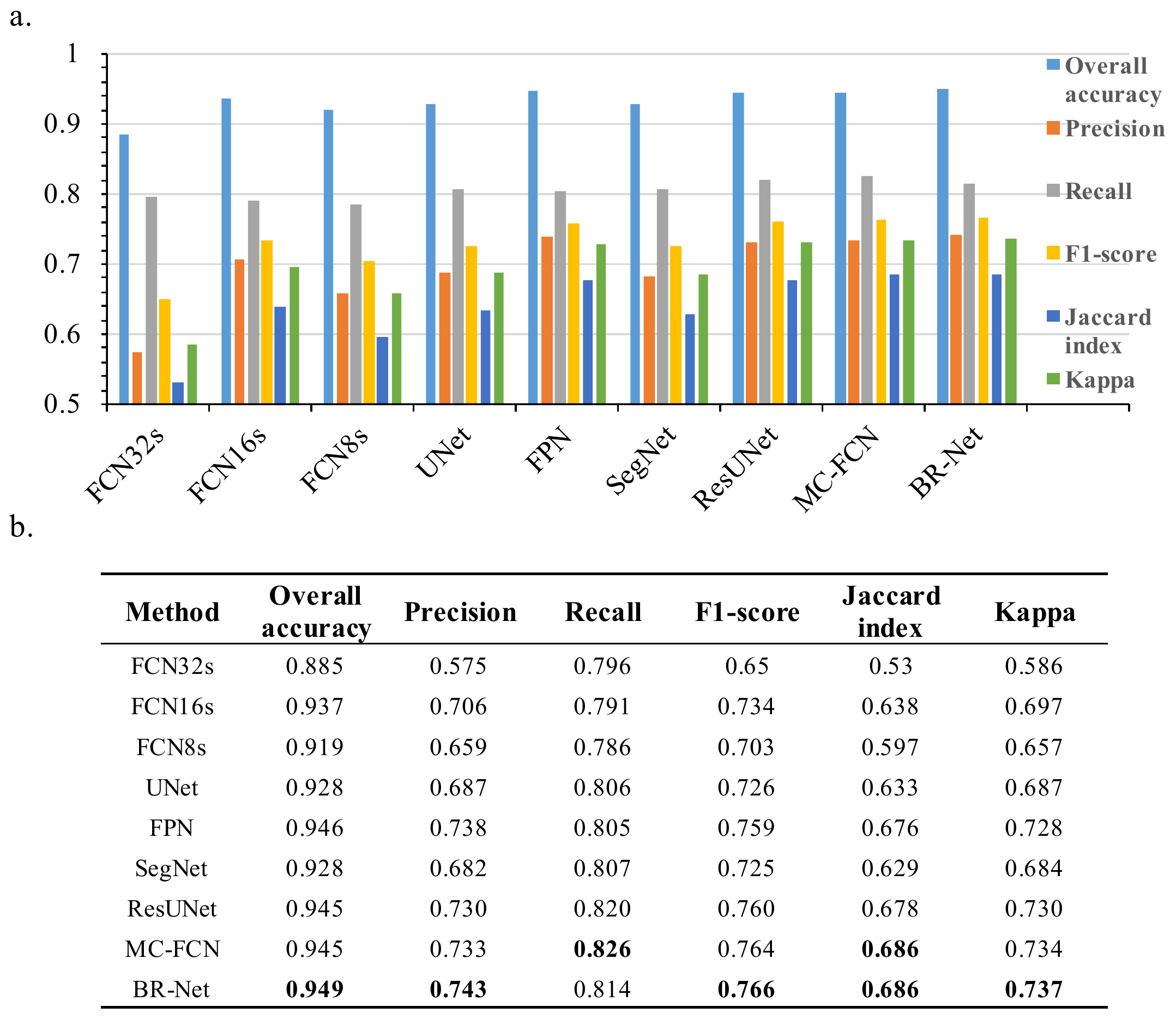}
\caption{Comparison of segmentation performances of implemented models across the entire testing data. (a) Bar chart for performance comparison. The x- and y-axis represent the implemented methods and corresponding performances, respectively. (b) Table of performance comparisons of methods. For each evaluation metric, the highest values are highlighted in \textbf{bold}.}
\label{fig:performance}
\end{figure}

For model evaluations, two imbalanced metrics of precision and recall, and four general metrics of overall accuracy, F1 score, Jaccard index, and kappa coefficient are utilized for quantitative evaluations. Figure \ref{fig:performance} presents comparative results between FCN8s, FCN16s, FCN32s, U-Net, FPN, ResUNet, MC-FCN and BR-Net for the testing samples.

For the imbalanced metrics of precision and recall, the BR-Net method achieves the highest value of precision (0.743) which indicates that the method performs well in terms of suppressing false positives. And, the MC-FCN method gains the highest value of recall (0.824) among nine implemented methods.

For the four general metrics,  the BR-Net model achieves the highest values for overall accuracy, F1 score, Jaccard index, and kappa coefficient.
Compared with the weakest model (FCN32s), the best model (BR-Net) achieves improvement of approximately 7.2\% (0.949 vs. 0.885) on overall accuracy. For F1 score, the best model achieves improvement of about 17.8\% (0.766 vs. 0.650) over FCN32s. Compared to the FCN32s method, the BR-Net method achieves improvements of 29.4\% (0.686 vs. 0.530) and 25.8\% (0.737 vs. 0.586) for Jaccard index and kappa coefficient, respectively. Considering the fact that all these models are proposed within three years, we can imagine the evolution speed within the research field.

\subsection{Computational efficiency} \label{subsec:efficiency}

The nine models are all implemented in PyTorch and tested on a 64-bit Ubuntu system equipped with an NVIDIA GeForce GTX 1070 GPU \footnote{\url{https://www.nvidia.com/en-us/geforce/products/10series/geforce-gtx-1070-ti/}}. During iterations, the Adam optimizer \cite{kingma2014adam} with a learning rate of 2e-4 and betas of (0.9, 0.999) is utilized. To ensure a fair comparison of the different methods, the batch size and iteration number for training are fixed as 24 and 5,000, respectively.

\begin{figure}[!hbt]
\centering
\includegraphics[width=\columnwidth]{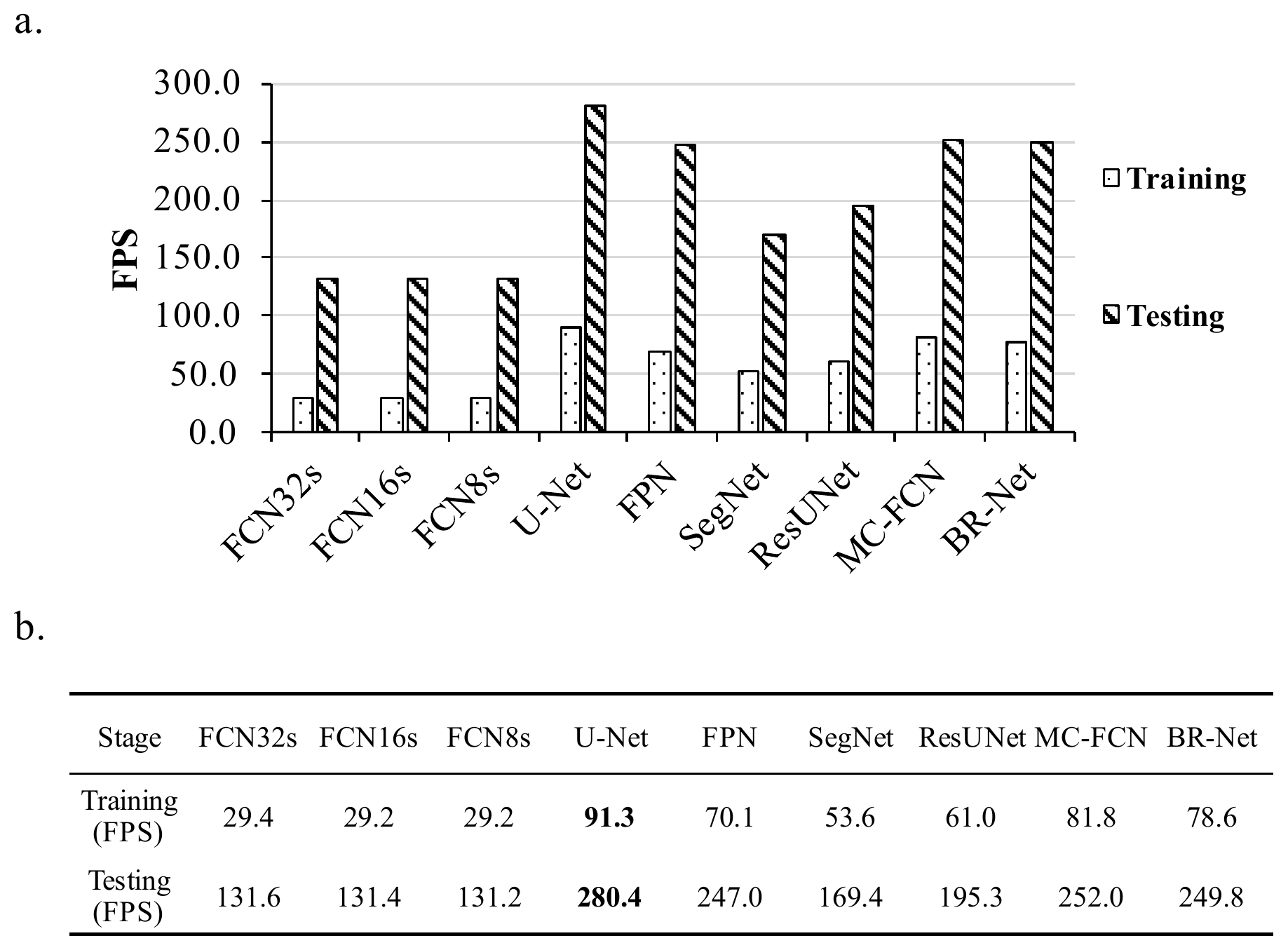}
\caption{Comparison of computational efficiency of the nine implemented methods.(a) Bar chart for computational efficiency comparison. The x- and y-axis represent the implemented methods and corresponding processing speed of frames per second (FPS), respectively. (b) Table of performance comparisons of methods. For each stage, the highest values are highlighted in \textbf{bold}.}
\label{fig:computation}
\end{figure}

The computational efficiencies of the different methods during different stages are listed in Figure \ref{fig:computation}. During the training stage, the slowest models (FCN8s and FCN16s) process approximately 29.2 FPS, while the fastest model (U-Net) reaches 91.3 FPS. Because of fewer computational operations, at the testing stage, the slowest model(FCN32) and the fastest model (U-Net) reach 131.6 and 280.4 FPS, respectively.

Even with slight differences in their architectures, three FCNs variants (FCN32s, FCN16s, and FCN8s) show almost identical computational efficiency at both training and testing stages. Consider the huge differences in their performances (see details in Figure \ref{fig:performance} b), it is better to avoid applying FCN32s model.

Compared with U-Net, more complex models such as FPN, ResUNet, MC-FCNthe and BR-Net adopt extra computation layers that lead to a slightly slower processing speed at both training and testing stages. The SegNet model, which is slower and weaker than U-Net, is also not a good option for robust building segmentation and outline extraction.

\section{Conclusion} \label{sec:conclusion}

In this paper, we introduce a computer vision package termed Geoseg that focus on accurate building segmentation and outline extraction. The Geoseg is built on top of PyTorch, a dynamic deep learning framework. In Geoseg, we implement nine models as well as utilities for handling dataset, logging, training, evaluating and visualization. Through a large-scale aerial image dataset, we evaluate performances and computational efficiency of implemented models including FCN32s, FCN16s, FCN8s, U-Net, FPN, SegNet, ResUNet, MC-FCN, and BR-Net. In comparison to the weakest model (FCN32s), the best model (BR-Net) achieves increments of 17.8\% (0.766 vs. 0.650), 29.4\% (0.686 vs. 0.530), and 25.8\% (0.737 vs. 0.586) in F1-score, Jaccard index, and kappa coefficient, respectively. In future studies, we will further optimize our network architecture to achieve better performance with less computational cost.

\section*{Acknowledgement}
We would like to thank SAKURA Internet Inc.(\url{https://www.sakura.ad.jp/}) for providing us the "koukaryoku" GPU server for testing.

\bibliographystyle{unsrt}
\bibliography{refs}

\end{document}